\documentclass[journal]{IEEEtran}
\ifCLASSINFOpdf
\else
\fi
\usepackage{amsmath}
\usepackage{amssymb}
\usepackage{mathtools}
\usepackage{balance}
\usepackage{algorithm}
\usepackage[noend]{algpseudocode}
\algnewcommand\algorithmicinput{\textbf{Input:}}
\algnewcommand\Input{\item[\algorithmicinput]}

\algnewcommand\algorithmicoutput{\textbf{Output:}}
\algnewcommand\Output{\item[\algorithmicoutput]}

\usepackage{amsthm} 

\usepackage{subcaption}

\usepackage{wrapfig,lipsum,booktabs}

\usepackage{capt-of}

\usepackage{multirow}

\usepackage{xcolor}
\begin{document}
	%
	\title{Adversarially Approximated Autoencoder for Image Generation and Manipulation}
	%
	%
	%
	\author{{Wenju~Xu,}
		Shawn Keshmiri,~\IEEEmembership{Member,~IEEE,}
		and~Guanghui~Wang,~\IEEEmembership{Senior Member,~IEEE}
    \thanks{The work was supported in part by NSF NRI and USDA NIFA under the award no. 2019-67021-28996, NSFC under grant No. 61573351, and the Nvidia GPU grant.}
	\thanks{Manuscript received XXXX, 2018; revised XXXX, 2018.}}

	\maketitle
	
	\begin{abstract}
		Regularized autoencoders learn the latent codes, a structure with the regularization under the distribution, which enables them the capability to infer the latent codes given observations and generate new samples given the codes. However, they are sometimes ambiguous as they tend to produce reconstructions that are not necessarily faithful reproduction of the inputs. The main reason is to enforce the learned latent code distribution to match a prior distribution while the true distribution remains unknown. To improve the reconstruction quality and learn the latent space a manifold structure, this work present a novel approach using the adversarially approximated autoencoder (AAAE) to investigate the latent codes with adversarial approximation. Instead of regularizing the latent codes by penalizing on the distance between the distributions of the model and the target, AAAE learns the autoencoder flexibly and approximates the latent space with a simpler generator. The ratio is estimated using generative adversarial network (GAN) to enforce the similarity of the distributions. Additionally, the image space is regularized with an additional adversarial regularizer. The proposed approach unifies two deep generative models for both latent space inference and diverse generation. The learning scheme is realized without regularization on the latent codes, which also encourages faithful reconstruction. Extensive validation experiments on four real-world datasets demonstrate the superior performance of AAAE. In comparison to the state-of-the-art approaches, AAAE generates samples with better quality and shares the properties of regularized autoencoder with a nice latent manifold structure.
	\end{abstract}
	

	%
	\IEEEpeerreviewmaketitle

	\section{Introduction}
	Image synthesis is to generate new samples that look as realistic as the real images \cite{reed2016generative,zhang2017stackgan}, or manipulate the images according to the objects' attributes \cite{upchurch2017deep,hou2018learning,liu2016deep}. This study has a large amount of potential applications, such as salient image generation \cite{wang2018video,wang2018deep,wang2018deep2}, high-resolution image generation \cite{johnson2016perceptual}, image style translation \cite{zhu2017unpaired, isola2017image}, domain adaption \cite{tzeng2017adversarial}, etc. The most promising solution to this problem is based on deep generative modes, which can be classified into three categories \cite{voulodimos2018deep}. The first one is the normalizing flows \cite{kingma2018glow,kingma2016improved,rezende2015variational,huang2018neural}, which learn maps between a simple distribution and a complex distribution through an invertible neural network. However, learning the invertible model is at a high cost of both the computational power and physical storage space. Thus, it is not a model for general computer vision tasks.
	
	The second category is the generative adversarial network (GAN) \cite{goodfellow2014generative}, which has made a significant advance towards development of generative models that are capable of synthesizing realistic data \cite{che2016mode,reed2016generative, zhang2017stackgan,salimans2017pixelcnn++}. The GAN takes random variables that are drawn from a simple explicit distribution as input and discourages the network from yielding unrealistic synthetic data. Thus, GAN yields a model capable of representing sophisticated distributions. Although the original GAN and its variants have been used to synthesize highly realistic data (e.g., images) \cite{radford2015unsupervised,denton2015deep,zhu2016generative,ledig2017photo}, those models lack the ability to infer the latent variables \cite{bengio2013representation,vincent2010stacked,donahue2016adversarial} given the observed
	data. This limitation has been mitigated using different approaches such as variational autoencoder (VAE) \cite{kingma2013auto} and other related approaches \cite{huszar2017variational,makhzani2015adversarial,tolstikhin2017wasserstein,pu2017adversarial,zhang2017age,gulrajani2016pixelvae}. In these methods a penalized form of distance between distributions of the model and target are minimized. They exploit a regularization term to encourage the encoded training distribution to match the prior distribution. Usually this prior is defined as a Gaussian distribution. 
	
	The regularized autoencoders have shown significant progress in learning smooth representations of complex, high dimensional data such as images. However, the regularized autoencoder are inadequate from the viewpoint of image reconstruction.
	Given the observed data and associated inferred latent
	variables, subsequently synthesized data often do not
	look particularly close to the original data. One possible reason for the difficulty in faithful reproductions is due to mismatches between the prior and true distributions. 
	Some effort has been taken to make the prior/posterior more flexible through the adversarial training. In the adversarial atuoencoder (AAE) \cite{makhzani2015adversarial} framework,
	a discriminator is trained to distinguish samples from a fixed prior distribution and the
	input encoding. The goal is to ensure code distributions matching the prior.
	Wasserstein Autoencoders (WAE) \cite{tolstikhin2017wasserstein} takes a weaker topology to measure the distance between two probability distributions. However, these works are still enforcing the latent distribution to match an explicit distribution, and thus degrade the quality of generations.
	
	In practice, unregularized autoencoders (UAE) often learn an identity mapping where the latent
	code space is learned to be the true distribution but free of any explicit structure. The popular approach is to regularize through an explicit prior on the code distribution and utilize a variational approximation to the posterior \cite{kingma2013auto,mescheder2017adversarial} or adversarial approximation to the posterior \cite{makhzani2015adversarial,tolstikhin2017wasserstein}. In this work, instead of enforcing the latent distribution to match a fixed explicit prior, we take the latent space learned by an unregularized autoencoder as the "prior". This latent space is then approximated through a simpler generator. Inspired by previous studies on UAEs and GANs \cite{kim2017adversarially,tran2017hierarchical,srivastava2017veegan}, we propose a novel unregularized adversarially approximated autoencoder (AAAE) to approximate the implicit probability distribution. In addition to the autoencoder, which learns an identity mapping, AAAE introduces an additional approximator network for mapping Gaussian random noise to the encoded latent code distribution. We train the autoencoder with an additional discriminator to approximate the data distribution. Learning a more accurate identity mapping function will encourage a true posterior distribution. Thus, the generated latent codes distribution by the approximator network is simultaneously matched to the posterior with adversarial training.
	
	The fundamental feature of our approach is to unify the AE and GAN in one framework. The encoder and generator constitute the AE, which tends to produce faithful reconstructions of the inputs. The approximator and encoder provides "fake" and "real" samples for the discriminator, which encourages the approximator to the map Gaussian random noise to the encoded latent code distribution. This also allows us to retrieve a reliable geometry of the latent manifold in both the encoded latent space and the generated latent space. The main contributions of this work are listed below.
	\begin{itemize}
		\item  A new structure of autoencoders, named as adversarially approximated autoencoder (AAAE), is proposed for image synthesis. To generate high-quality samples and manipulate the observed images, two types of objectives are jointly optimized: the reconstruction cost $c(x,G_{\psi})$ between the data $x$ and the decoded image $G_{\psi}$, and the cross entropy $H(G_{\phi},p_{\theta})$ indicating the discrepancy between these two distributions.
		
		\item The adversarial training scheme combined with the cross entropy is proposed to estimate the implicit probability distributions. They jointly encourage to learn the data space and the latent code space. Instead of matching the latent code distribution to a prior explicit distribution, we approximate the learned latent code distribution by employing two different adversarial learning schemes.
		
		\item The proposed neural network architecture contains one encoder-generator pair, one approximator and two discriminators. In our proposed approach, $ p_{\theta}$ is the encoded latent code distribution from the encoder, while $G_{\phi}$ is the generated latent code distribution from the approximator $G_{\phi}(z)$ given $z \sim N (0, I)$. This architecture promotes the adversarial regularizations under both the data space and the latent code space.
	\end{itemize} 
	
	The proposed AAAE model is extensively evaluated and compared on four public datasets, MNIST, CIFAR-10, CelebA and Oxford-102 datasets, in terms of faithful reconstruction and generation quality. Both quantitative and qualitative results demonstrate the superior performance of AAAE compared with the state-of-the-art approaches.
	
	The remainder of this paper is organized as follows. We briefly review the related works and provide some background in Section \ref{background}. In Section \ref{AAAE}, our proposed AAAE approach is explained in detail. Then Section \ref{experiments} demonstrates the experimental results as well as analyses. Finally, this paper is concluded in Section \ref{conclusion}.
	
	\section{Background and Related Work} \label{background}
	\subsection{Deep Probabilistic Model}
	In recent years, a lot of attention has been paid in developing deep models with data from a large quantity, and offering the unlabeled data. Deep neural network models have shown great potential in dealing with various computer vision tasks, such as image generation \cite{dumoulin2016adversarially,tolstikhin2017wasserstein}, image translation \cite{zhu2017toward,larsen2016autoencoding}, object detection and recognition \cite{ma2018mdcn, gao2016novel}, super-resolution imaging \cite{johnson2016perceptual,ledig2016photo}, face image synthesis \cite{tran2017DRGAN,zhang2017age}, depth estimation \cite{he2018learning, he2018spindle}, and image representation \cite{cui2018general}. Through the learning frame, we can build rich and hierarchical probabilistic models to fit the distribution of complex real data and synthesize realistic image data \cite{mathieu2016disentangling,chen2016infogan}. 
	
	Given the data sample $x\sim p_x$, where $p_x:=p(x)$ is the true and unknown distribution, and $x \in \{x_i\}_{i=1}^N$ is the observed data for training, the purpose of generative model is to fit the data samples with the model parameters $\psi$ and random code $z$ sampled from an explicit prior distribution $p_z:=p(z)$. This process is denoted by $x \sim q_{\psi}(x|z)$ and the training is to model a neural network $G_{\psi}$ that maps the representation vectors z to data x. For inference, the true posterior distribution $p(z|x)$ is intractable. We approximate this distribution with $p_{\theta}(z|x)$, which is the output of a neural network $E_{\theta}$ with the learned parameters $\theta$. Two types of deep probabilistic models, GAN and VAE, are distinguishing among recent researches on deep probabilistic models. They both learn to approximate the data distribution, while VAE employs an additional neural network $E_{\theta}$ for inference. The following parts will portray these two models in details.
	
	\subsection{Adversarial Approximation to Implicit Probability Distribution}
	\label{app}
	The generative adversarial network (GAN) is built on the map function $p_{\theta}(x|z)$ without an inference network $q_{\phi}(z|x)$. To approximate the true distribution $p(x)$, the model is trained by introducing a second neural network discriminator. 
	\begin{equation*}
	\begin{aligned}
	D_{GAN}(P_X,P_G) &= \mathbb{E}_{x \sim P_x}[\log D(X)] \\&+ \mathbb{E}_{z \sim P_z}[\log(1-D(G(z))]
	\end{aligned}
	\end{equation*}  
	The discriminator $D$ is trained to distinguish between the real and the generated samples, and the generator is employed to confuse the discriminator with respect to a deterministic decoder $G_{\psi}: \mathcal{Z} \to \mathcal{X}$. The discriminator can provide us a measure on how probable the generated sample is from the true data distribution. WGAN \cite{arjovsky2017wasserstein, gulrajani2017improved} is trained with Wasserstein-1 distance to strengthen the measure on the probability divergence and thus improves the training stability. While the GANs do not allow inference of the latent code. To solve this issue, BEGAN \cite{berthelot2017began} applies an auto-encoder
	as the discriminator. ALI \cite{dumoulin2016adversarially} proposed to
	match on an augmented space (variational
	joint $p(x, z) = p(x)p(z | x)$), which simultaneously trains the model and an inverse mapping from
	the random noise to data. A recent successful extension, VEEGAN \cite{srivastava2017veegan}, is also trained by discriminating the joint samples of the data and the corresponding latent variable $z$, by introducing an additional regularization to penalize the cross entropy of the inferred latent code. To approximate to the augmented space, the objective function for the discriminator is written as 
	
	\begin{equation*}
	\begin{aligned}
	D(\theta, \phi) &= \mathbb{E}_{x\sim q_{\psi},z\sim p_z}[\log \sigma(r(x,z))] \\&+ \mathbb{E}_{x\sim p_x, z\sim p_{\theta} }[\log(1-\sigma(r(x,z))]
	\end{aligned}
	\end{equation*}
	where $\sigma$ is the sigmoid function, and $q_{\psi}$ and $p_{\theta}$ are the outputs of the neural network. According to ratio estimation \cite{tran2017hierarchical}, the discriminator is to estimate the ratio 
	\begin{equation*}
	\begin{aligned}
	r(\theta, \phi) = \log q_{\theta,\psi}(x,z)-\log p(x,z)
	\end{aligned}
	\end{equation*}
	where $q_{\theta,\psi}(x, z)$ is the output of the stochastic neural network $G_{\psi}$ and $E_{\theta}$. This framework adversarially estimate the ratio of the two distributions instead of the probability divergence. In case of implicit representations for $q_{\phi}$ and $p_{\theta}$, the ratio estimation is an ideal alternative for the probability distances and divergences, which is the key idea of GANs. 
	
	\subsection{Regularized Autoencoder}
	Another type of generative model is the regularized autoencoder, which learns the latent code space a nice manifold structure. Given an explicit prior distribution $p(z)$, the generative modeling process is to sample $x \sim p(x|z)$. However, direct evaluation of $p(x)=\int p(x|z)p(z)dz$ is typically intractable. One popular approach is to introduce a regularizer to jointly optimize the $KL$ divergence $D_{KL}(p_{\theta}(z|x),p_z)$ and the mutual information between $x$ and the latent codes $z$, which is successfully employed by the variational auto-encoders (VAE) \cite{kingma2013auto}. It utilizes variational approximation and minimizes
	\begin{equation*}
	\begin{aligned}
	D_{VAE}(P_X,P_G) &= \inf_{Q(Z|X)\in Q} \mathbb{E}_{ P_x}[D_{KL}(Q(Z|X),P_Z) \\&- \mathbb{E}_{Q(Z|X)}[log(P_G{X|Z})]]
	\end{aligned}
	\end{equation*}  
	
	The VAE formulation depends on a random encoder mapping function $G_{\theta}$, and take a "reparametrization trick" to optimize the parameter $\theta$. Moreover, minimizing the $KL$ divergence drives the $p_{\theta}(z|x=x_i)$ to match the prior $p(z)$, thus the solution will converge around the optima. One possible extension is to force the mixture $p_{\theta}:=\int p_{\theta}(z|x)dp_x $ to match the prior. With this observation, AAE \cite{makhzani2015adversarial} and WAE \cite{tolstikhin2017wasserstein} regularize the latent code space with the adversarial training. WAE minimizes a relaxed optimal transport by penalizing the divergence between $p_{\theta}$ and $p_z$
	
	\begin{equation*}
	\begin{aligned}
	D_{WAE}(p_x,q_{\psi}) &=\inf_{p(z|x)\in p_z}\mathbb{E}_{ P_x}\mathbb{E}_{p(z|x)}[c(x,G_{\phi}(z))] \\& + \lambda D_{z}(p_{\theta},p_z) 
	\end{aligned}
	\end{equation*} 
	
	This formulation attempts to match the encoded distribution of the training examples $p_{\theta} = \mathbb{E}_{p_x} [p(z|x)]$ to the prior $p_z$ as measured by any specified divergence
	$D_{z}(p_{\theta},p_z)$ in order to guarantee that the latent codes provided to the decoder are informative
	enough to reconstruct the encoded training examples. It also allows the non-random encoders deterministically map the inputs to their latent codes. This gives raise to the potential of unifying two types of generative models \cite{hu2017unifying,mescheder2017adversarial,larsen2016autoencoding,makhzani2017pixelgan} in one framework. 
	
	Inspired by this idea, the proposed model is built on the classical auto-encoders. It jointly minimizes a reconstruction cost and an adversarial loss in the image space while there is no explicit regularization on the latent code space. The encoder $E_{\theta}$ trained in this way will provide an efficient representation $c$, and the samples from the decoder $G_{\psi}(c)$ will conform to the true data distribution. This results in different training points being encoded into a nice latent manifold structure without "holes" in between. Thus, we can build a simpler approximator to generate codes that exactly lay in the encoded latent space $p_{\theta}$. In next section, we will discuss this process in details.
	
	\section{Autoencoder with Adversarial Approximation} \label{AAAE}
\begin{figure*}[t]
		\begin{center}		
			\includegraphics[width=0.9\textwidth]{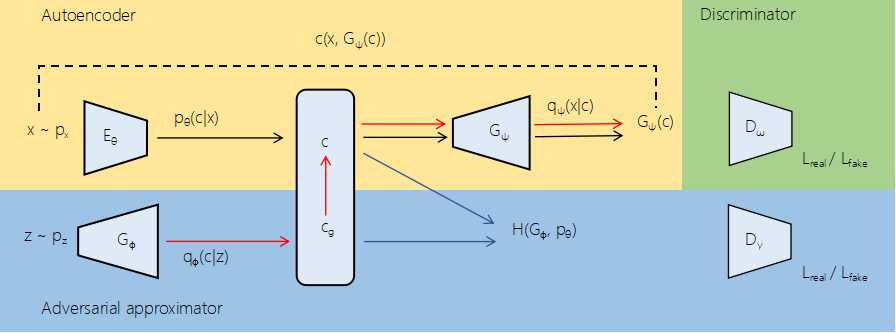}
		\end{center}
		\caption{An overview of the AAAE architecture. Black arrows denote the encoder-decoder structure; red arrows stand for the sampling process; and blue arrows indicate the constraint on the latent codes.}
		\label{f_AAAE}
	\end{figure*}

	To build an autoencoder for faithful reconstruction with a nice latent manifold structure, we propose to learn the autoencoder without the regularization on the latent codes distribution. The proposed AAAE consists of two major components: The autoencoder and the latent space approximator, as illustrated in Figure \ref{f_AAAE}.
	We train the deterministic encoder $E_{\theta}(c|x)$ and the conditional decoder $G_{\psi}(x|c)$ with a discriminator $D_{\omega}$. To this end, we assume that the successfully trained autoencoder ensures the $p_{\theta}(c|x)$ is the true latent codes distribution with an unknown structure, which cannot be sampled in a closed form. To solve the sampling issue, instead of regularizing the distribution $p_{\theta}(c|x)$ to match the prior $p_z$, we introduce a simpler generator $G_{\phi}$ to approximate the true latent codes distribution as $p_{\phi}(c|z) \subseteq p_c$ given randomly sampled noise $z\sim \mathcal{N}(0,I)$. We take the cross entropy $H(G_{\phi},p_{\theta})$ \cite{li2017alice,srivastava2017veegan} to quantify the divergence between the two probability distributions. The joint objective is defined as 
	
	\begin{align}
	\begin{aligned}
	\mathcal{O}_{AAAE}(\theta, \phi,\psi)&:= \inf_{p_{\theta}(c|x)\in p_c} \mathbb{E}_{ P_x}\mathbb{E}_{p(c|x)}[c(x,G_{\psi})] \\&+ H(G_{\phi},p_{\theta})   
	\end{aligned}
	\label{objective}
	\end{align}
	where $p_c$ is any output of a nonparametric encoder and $p_{\theta}$ is parameterized as the output of the neural network $E_{\theta}(c|x)$. This objective is not easy to solve. Next, we discuss how to simplify and transform it into a computable version.

	\begin{algorithm*}
		\caption{The training pipeline of AAAE.
		}\label{algo}
		\begin{algorithmic}[1]
			\While {not converged}
			\For {$i\in \{1...N\}$}
			\State Sample $x^i \sim p(x)$; Sample $x^i_g \sim q_{\psi}(x|c_i)$; Sample $c^i \sim q_{\theta}(c|x_i)$.
			
			\State
			\State $g_{\theta,\psi}\gets  \mathcal{r}   \frac{1}{N}\sum_{i}D_{\omega}(x^i) + \frac{\lambda_1}{N}\sum_{i}d(x^i,G_{\psi}(c^i))$ 	\hspace*{0pt}\hfill $\rhd$ Compute~$\mathcal{r}_{\theta,\psi}\mathcal{O}_{x}$  via Eq. (\ref{Ox})
			\State $g_{\omega}\gets  -\mathcal{r}  \frac{1}{N}\sum_{i}\log \sigma (D_{\omega}(x^i))+\log (1-\sigma (D_{\omega}(x^i_g)))$ \hspace*{0pt}\hfill $\rhd$ Compute~$\mathcal{r}_{\omega}\mathcal{O}_{x}$  via Eq. (\ref{Ox})	
			\State $\theta \gets \theta -\eta g_{\theta};\psi \gets \psi -\eta g_{\psi};\omega \gets \omega -\eta g_{\omega}$ \hspace*{0pt}\hfill $\rhd$ Perform SGD updates for $\theta,\psi$ and $\omega$		
			\State 	
			\For {$j\in \{1...k\}$}	
			\State Sample $z^j \sim p(z)$; Sample $c^j_g \sim q_{\phi}(c|z_j)$; Sample $x \sim p(x)$; Sample $c^j \sim q_{\theta}(c|x)$.				
			\State
			\State $g_{\gamma}\gets  \frac{1}{N}\sum_{i}[-D_{\gamma}(c^j)+D_{\gamma}(c_g^j)  +\lambda_2(||\triangledown D_{\gamma}(c_g^j)||_2-1)^2]$  \hspace*{0pt}\hfill $\rhd$ Compute~$\mathcal{r}_{\gamma}\mathcal{O}_{z} $  via Eq. (\ref{Oz}) 
			\State $g_{\phi}\gets  \frac{1}{N}\sum_{i}[-D_{\gamma}(c_g^j)] $  \hspace*{0pt}\hfill $\rhd$ Compute~$\mathcal{r}_{\phi}\mathcal{O}_{z} $   via Eq. (\ref{Oz})
			\State $\phi \gets \phi -\eta g_{\phi};\gamma \gets \gamma -\eta g_{\gamma}$ \hspace*{0pt}\hfill $\rhd$ Perform SGD updates for $\phi$ and $\gamma$	
			\EndFor
			\EndFor	 	
			\EndWhile
		\end{algorithmic}
	\end{algorithm*}
	
	\subsection{Objective Function}
	The first term in Eq. (\ref{objective}) is the optimal transport cost $W_c(p_x,G_{\psi})$ discussed in \cite{tolstikhin2017wasserstein}. Unlike the previous work which relaxes the constraints on the latent code by regularizing the encoded code to match a prior distribution, our formulation is an unregularized autoencoder, which learns an identity function and encourages $p_{\phi}(c|z)$ to be the true latent code distribution $p_c$. Since the autoencoder is trained without regularization on the latent codes, the adversarial training schedule, implemented by the conditional GANs \cite{isola2017image,zhu2017unpaired} is stable and improves the sharpness of the generations. 
	
	Let us denote $p_{\phi}(c|z) \subseteq p_c$, which is the distribution of the outputs of the approximator $G_{\phi}$, the cross entropy $H(G_{\phi},p_{\theta})$ is written as
	\begin{align}
	\begin{aligned}
	H(G_{\phi},p_{\theta})  = -\int p_{\phi}(c_g|z)\log p_{\theta}(c|x)dc_g
	\end{aligned}
	\end{align}
	
	To solve the integral numerically, we introduce two distributions $q_{\psi}(x|c)$ and $p_{\phi}(c_g|z)$. Here, we denote $c_g \sim p_{\phi}(c|z)$ as the generated latent code, which is to approximate the encoded latent code $c \sim p_{\theta}(c|x)$, and by Jensen's inequality, we have
	\begin{align}
	\begin{aligned}
	-\log p_{\theta}(c|x) &=  -\log\int p_{\theta}(c|x)p(x)\frac{q_{\psi}(x|c)p_{\phi}(c_g|z)}{q_{\psi}(x|c)p_{\phi}(c_g|z)}dx \\
	&\leq   -\int q_{\psi}(x|c)\log\frac{p_{\theta}(c|x)p(x)}{q_{\psi}(x|c)p_{\phi}(c_g|z)} dx \\&- \log p_{\phi}(c_g|z)
	\end{aligned}
	\end{align}
	which is used to bound the cross entropy. The $q_{\psi}(x|c)$ is parameterized with the decoder, enabling us to learn complex distribution. In practice, minimizing this bound is difficult if the distributions are implicit. In order to obtain a numerical solution, we write (see the supplementary document for proof)
	\begin{align}
	\begin{aligned}
	-\int p_{\phi}(c_g|z)\log p_{\theta}(c|x)dc_g &\leq KL[q_{\psi}(x|c)||p(x)] \\&+ KL[p_{\phi}(c_g|z)||p_{\theta}(c|x)] \\&-\mathbb{E}[\log p_{\phi}(c_g|z)]
	\end{aligned}
	\end{align}
	
	Thus, the final objective function can be written as
	\begin{align}
	\begin{aligned}
	\mathcal{O}(\theta,\phi,\psi) &= KL[q_{\psi}(x|c)||p(x)] + KL[p_{\phi}(c_g|z)||p_{\theta}(c|x)] \\&-\mathbb{E}[\log p_{\phi}(c_g|z)] + \mathbb{E}[c(p_x,G_{\psi})]
	\end{aligned}
	\label{kl}
	\end{align}
	
	Let us use $\mathcal{O}_x$ to denote the joint constraints of $KL[q_{\psi}(x|c)||p(x)] + \mathbb{E}[c(p_x,G_{\psi})]$, which is enforced to match the true data distribution, and use $\mathcal{O}_c$ to denote the combination of $KL[p_{\phi}(c_g|z)||p_{\theta}(c|x)] -\mathbb{E}[\log p_{\phi}(c_g|z)]$ that enforces the approximator to generate true codes. Our goal is to the minimize the upper bound with respect to $\theta$, $\phi$ and $\psi$. By finding the global  minimum of $\mathcal{O}$, it is guaranteed that $q_{\psi}(x|c)$ matches the true data distribution, while $p_{\phi}(c_g|z)$ follows the true latent codes distribution.
	
	\begin{table*}
		\centering
		{
			\renewcommand{\arraystretch}{1.5}
			\centering
			\caption{Quantitative results on real-world datasets. To compare the quality of random samples, we report ICP scores (higher is better) on MNIST and CIFAR-10 data and FID scores (smaller is better) on CelebA and Oxford-102 datasets. For the reconstruction quality, we report MSE (smaller is better). $\dag$ is the best performance reported in \cite{li2017alice}; $\ddag$ is calculated using the method in \cite{li2017alice}.}
			\label{t1}
			\begin{tabular}{c|cc|cc|cc|c}
				
				\hline
				\multirow{2}{*}{Settings} & \multicolumn{2}{|c|}{MNIST}&\multicolumn{2}{|c|}{CIFAR-10} &\multicolumn{2}{|c}{ CelebA} &\multicolumn{1}{|c}{ Oxford-102}\\ 
				\cline{2-8}
				& ICP&MSE& ICP&MSE& FID&MSE& FID\\ 
				\hline
				WGAN-GP \cite{gulrajani2017improved} & &&6.610& &&&\\ 
				ALI \cite{dumoulin2016adversarially} &  $8.839^{\dag}$ &$0.380^{\dag}$&$5.974^{\dag}$ &$0.559^{\dag}$&6.946&0.283&80.817\\ 
				ALICE \cite{li2017alice} & $9.349^{\dag}$&$0.073^{\dag}$&$6.043^{\dag}$ &$0.216^{\dag}$ &&&\\
				WAE \cite{tolstikhin2017wasserstein} &    &&&&98.780&0.019&196.560\\ 
				AAAE &$9.873^{\ddag}$  &0.011&6.431&0.053 &4.868&0.023&103.460\\ 
				\hline		
			\end{tabular}
		}
	\end{table*}
	
	\begin{figure*}
		\centering
		\begin{subfigure}[t]{0.32\textwidth}
			\centering
			\includegraphics[height=5cm]{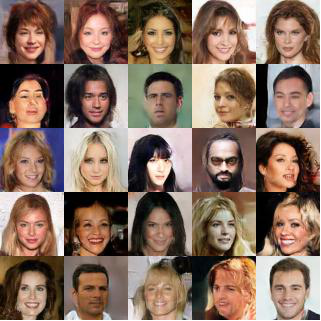}
			\caption{\label{celeba_AAAE} AAAE samples}
		\end{subfigure}
		\begin{subfigure}[t]{0.32\textwidth}
			\centering
			\includegraphics[height=5cm]{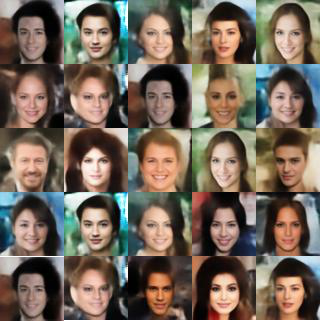}
			\caption{\label{celeba_WAE} WAE samples}
		\end{subfigure}
		\begin{subfigure}[t]{0.32\textwidth}
			\centering
			\includegraphics[height=5cm]{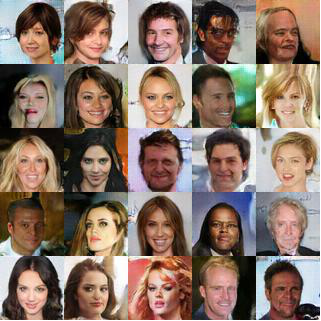}
			\caption{\label{celeba_ALI}  ALI samples}
		\end{subfigure}
		\caption{\label{celeba}Generated samples by different approaches trained on the CelebA dataset.}	
	\end{figure*}
	
	\begin{figure*}[ht!]
		\centering
		\begin{subfigure}[t]{0.3\textwidth}
			\centering
			\includegraphics[height=4.8cm]{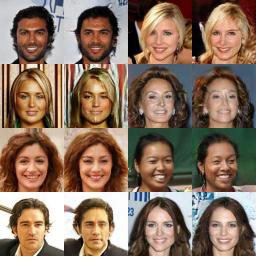}
			\caption{\label{celeba_recon_AAAE} AAAE reconstructions}
		\end{subfigure}
		\begin{subfigure}[t]{0.3\textwidth}
			\centering
			\includegraphics[height=4.8cm]{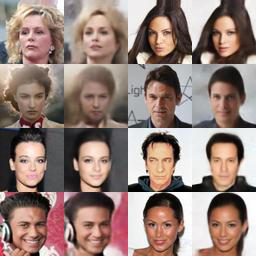}
			\caption{\label{celeba_recon_WAE} WAE reconstructions}
		\end{subfigure}
		\begin{subfigure}[t]{0.3\textwidth}
			\centering
			\includegraphics[height=4.8cm]{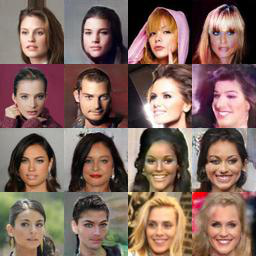}
			\caption{\label{celeba_recon_ALI}  ALI reconstructions}
		\end{subfigure}
		\caption{\label{celeba_recon} Comparison of the reconstructions on the CelebA dataset. In each block, odd columns represent the ground-truth, even columns are the reconstructions.}	
	\end{figure*}
	
		\begin{figure*}[h!]
		\begin{center}
			{\includegraphics[height=10cm]{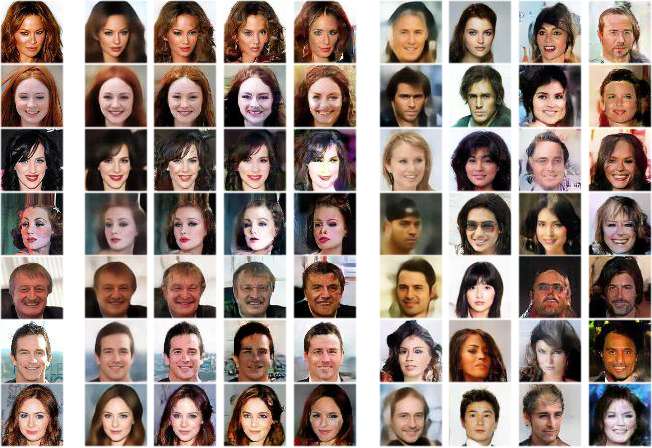}}
			
		\end{center}
		\caption{CelebA results with different $\lambda_1$. Left block: real inputs. Middle block: AAAE reconstructions. Right block: AAAE samples. $\lambda_1 = 0.0001, 0.001, 0.01$ and $0.1$ from left to right in each block.
		}
		\label{comp}
	\end{figure*}
	\subsection{Approximation to Implicit Probability Distribution}
	In this subsection, we apply ratio estimation to approximate the implicit probability distribution.
	The KL divergences in Eq. (\ref{kl}) depend on the density ratio of the two intractable distributions. In this case, we cannot optimize $\mathcal{O}$ directly. Following the discussion in Section \ref{app}, we estimate these ratios using two separate discriminators. We first minimize the $\mathcal{O}_x$ using the discriminator network $D_{\gamma}(x)$, which is trained to encourage 
	\begin{align}
	\begin{aligned}
	D_{\omega}(x) = \log q_{\psi}(x|c) -\log p(x)
	\end{aligned}
	\end{align}
	
	This will lead to the loss function, 
	\begin{align}
	\begin{aligned}
	\mathcal{O}_{x}(\theta,\psi,\omega) =  \frac{1}{N}\sum_{i=1}^{N}D_{\omega}(x^i) + \frac{\lambda_1}{N}\sum_{i=1}^{N}d(x,G_{\psi}(c^i))
	\end{aligned}
	\label{Ox}
	\end{align}
	where $D_{\omega}(x)$ makes use of the key idea of patch discriminator \cite{isola2017image,zhu2017unpaired} to discriminate the image patches instead of the whole image.
	
	To approximate to the latent space, we minimize $\mathcal{O}_z$ with another discriminator network $D_{\gamma}(c)$ to distinguish the latent codes, 
	\begin{align}
	\begin{aligned}
	D_{\gamma}(c) = \log p_{\phi}(c_g|z) -\log p_{\theta}(c|x)
	\end{aligned}
	\end{align}
	
	Note that, within the $\mathcal{O}_z$, the $-\mathbb{E}[\log p_{\phi}(c_g|z)]$ requires the $p_{\phi}(c_g|z)$ to match the true latent distribution. We optimize the $L$1 Wasserstein distance \cite{arjovsky2017wasserstein,gulrajani2017improved}
	\begin{align}
	\begin{aligned}
	\mathcal{O}_{z}(\phi,\gamma) &=  \frac{1}{N}\sum_{i=1}^{N}[-D_{\gamma}(c^i)+D_{\gamma}(c_g^i)  \\&+\lambda_2(||\triangledown D_{\gamma}(c_g^i)||_2-1)^2]
	\end{aligned}
	\label{Oz}
	\end{align}
	where $c_g \sim p_{\phi}(c|z)$ and $c \sim p_{\theta}(c|x)$.
	We assume $p_{\phi}(c|z) \subseteq p_c$. In the training process, the $p_{\phi}(c|z)$ is gradually approximating $p_c$. A practical way is to enforce the $p_{\phi}(c|z)$ to approximate $p_{\theta}(c|x)$.  With a well trained unregularized autoencoder, $p_{\theta}(c|x)$ follows in the true latent codes distribution, We can gradually enforce $p_{\phi}(c|z)$ to match the $p_{\theta}(c|x)$ so as to match the true latent code distribution and $-\mathbb{E}[\log p_{\phi}(c|z)]$ is minimized. In practice, we always train the autoencoder first then train the approximator adversarially for multiple iterations with the 'real' latent codes provided by the autoencoder. This training procedure is described in Algorithm \ref{algo}.
	
	\begin{table}[t!]
	\renewcommand{\arraystretch}{1.2}
	\renewcommand{\tabcolsep}{0.5cm}
	\centering
	\caption{The effect of discriminator $D_{\omega}$. AAAE CelebA quantitative evaluation with different values of $\lambda_1$}.
	\label{t22}	
	\vspace{2.0mm}
	\begin{tabular}{c|cccc}
		\hline
		$\lambda_1$&0.0001&0.001&0.01&0.1\\
		\hline
		\hline
		FID &  88.564&4.868&18.541&22.245\\
		MSE   &0.018 &0.023 &0.033&0.039\\
		\hline
	\end{tabular}
\end{table}
	
	\begin{figure*}[ht!]
		\centering
		\hspace*{\fill}
		\begin{minipage}[t]{0.49\textwidth}
			\centering
			\vspace{0pt}
			\includegraphics[height=5.1cm]{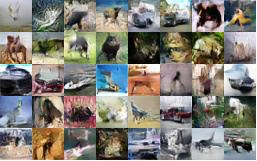}
			\caption{\label{cifar_samples} Samples generated by AAAE
				when trained on CIFAR-10.}
		\end{minipage}
		\hfill
		\hspace*{\fill}
		\begin{minipage}[t]{0.49\textwidth}
			\centering
			\vspace{0pt}\raggedright
			\includegraphics[height=5.1cm]{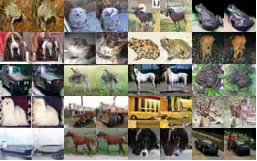}
			\caption{\label{cifar_reconst} Reconstructions generated by AAAE on CIFAR-10 (odd columns are ground-truth, and even columns are reconstructions).}
		\end{minipage}
	\end{figure*}
	
	\begin{figure*}[ht!]
		\centering
		\includegraphics[height=1.4cm]{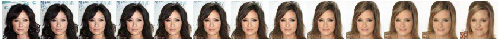}
		\caption{\label{inte}Latent space interpolations. The left and right columns are the original data pairs, and, the columns in between are the reconstructions generated with linearly interpolated latent codes.}	
	\end{figure*}
	
	\begin{figure*}[ht!]
		\centering
		\includegraphics[height=1.95cm]{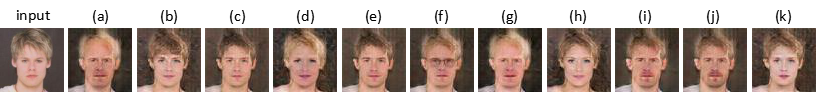}
		\caption{\label{attributes}Using AAAE to reconstruct samples with manipulated latent codes. (a) bald; (b) bangs; (c) black hair; (d) blond hair; (e) bushy eyebrows; (f) eyeglasses; (g) gray hair; (h) heavy makeup; (i) male; (j) mustache and (k) pale skin.}
	\end{figure*}
	
	\section{Experiments} \label{experiments} 
	In this section, we evaluate the proposed AAAE model and compare its performance quantitatively and qualitatively with the state-of-the-art approaches. Four publically available datasets are used to train the model: MNIST consisting of 70k images, CIFAR-10 \cite{krizhevsky2009learning} consisting of 60k images in 10 classes, CelebA \cite{liu2015deep} containing roughly 203k images and Oxford-102 \cite{nilsback2008automated} containing about 9k images of flowers. We report our results on three aspects of the model. First, we measure the reconstruction accuracy of the data points and the quality of the randomly generated samples. Next, we explore the latent space by manipulating the codes for consistent image transformation. Finally, we employ the learned generator network as an implicit latent variable model over complicate distribution. The model should generalize well and be robust to model collapsing.

	\textbf{Implementation details:}
	 We employ Adam \cite{kingma2014adam} to optimize our model with the learning rate of 0.0001. For the CelebA dataset, we crop the original images from $178\times218$ to $178\times178$ centered at the faces, then resize them to $64\times64$. For the Oxford-102 data, we resize the initial images to $64\times64$ without cropping. The network architectures of encoder, approximator and discriminator are provided in the supplementary document.
	
	We conduct experiments on generating high-quality synthesized images. At the training stage, we learn the autoencoder model and utilize $k = 2$ steps to update the approximator model within each learning iteration. We set $\lambda_1 = 0.001$, $\lambda_2 = 10$, and do not perform any dataset-specific tuning except for employing early stopping based on the average data reconstruction loss of $x$ on the validation sets. At the testing stage, the new samples are generated as $x_g = G_{\psi}(G_{\theta}(z))$ with $z$ sampled from $\mathcal{N}(0,I)$.
	
	\textbf{Quantitative evaluation protocol:} We adopt mean squared error (MSE) and inception score (ICP) \cite{salimans2016improved,li2017alice} to quantitatively evaluate the performance of the generative models. MSE is employed to evaluate the reconstruction quality, while ICP reflects the plausibility and variety of the sample generation. Lower MSE and higher ICP indicate better results. In order to quantitatively assess the quality of the generated images on the CelebA and Oxford-102 datasets, we adopt the Frechet inception distance (FID) introduced in \cite{heusel2017gans}. The ICP and FID scores are computed statistically based on $10,000$ samples.  
	
	
	\subsection{Random Samples and Reconstruction}
	Random samples are generated by sampling $p_z$ and generating the code $c_g\sim q_{\phi}(c_g|z)$, and decoding the resulted code vectors $c$ into $G_{\psi}(c)$. As expected, we observed in our experiments that the quality of samples strongly depends on how accurately $q_{\phi}(c_g|z)$ matches $p_c$. Note that while training the decoder function, $G_{\psi}$ is presented only with encoded versions $p_{\theta}(c|x)$ of the data points $x \sim p_x$ and there is no reason to expect good results without feeding it with samples from $p_c$. Indeed, the decoder is adversarially trained with samples from $p_x$ and thus it is reasonable to expect good results on the encoded codes. In our experiments we notice that even slight differences between $p_{\theta}(c|x)$ and $p_c$ may affect the quality of the samples. In the experiments, AAAE is demonstrated to lead to a better matching and generates better samples. It is robust to the mode collapse, which happens when the samples from $q_{\psi}(x|c)$ capture only a few of the modes of $p(x)$. 
	
	The quantitative results are tabulated in Table \ref{t1}. It is evident that AAAE not only captures the most modes, it also consistently matches the data distribution more closely than all other methods. As shown in the table, our model achieves the lowest MSE and highest ICP scores, and it outperforms its counterparts: ALI, WAE and ALICE. Our model achieves compelling results on both reconstruction and generation. Compared with the state-of-the-art models, AAAE yields competitive results on the randomly sampling, and a much better performance on the reconstruction in comparison with other deep generative models. AAAE achieve the lowest MSE (0.011) and highest ICP (9.873) on MNIST. In the experiments, we note that, for high dimensional data like the CelebA, ALI yields a very high MSE, while AAAE achieves a much better performance in terms of both FID and MSE. 
	
	The generated samples are shown in Figure \ref{celeba} and the reconstructions are listed in Figure \ref{celeba_recon}. It is evident that the reconstructions of ALI are not faithful reproduction of the input data, although they are related to the input images. The results demonstrate the limitation of adversarial training in reconstruction. This is also consistent with the results in terms of MSE as shown in Table \ref{t1}. In Figure \ref{comp}, We compare the different generations from the models trained with different values of $\lambda_1$. In table \ref{t22} we show quantitative results in terms of FID score and mean squared error of AAAE with different values of $\lambda_1$. As can be seen, the model is able to reconstruct images when $\lambda_1$ increases. However, when $\lambda_1$ is larger than 0.001, we observe an increase in the FID score and a decrease in the MSE. The small value of $\lambda_1$ leads to a better MSE but blurring generations, while bigger value of $\lambda_1$ yields stronger discriminator, which degrades the generator. The experimental results prove that discriminator $D_{\omega}$ plays a key role on the visual quality of both random sampling and faithful reconstruction. Figures \ref{cifar_samples} and \ref{cifar_reconst} illustrate the results of image generation and reconstruction on the CIFAR-10 dataset, respectively. More experimental results are provided in the supplementary document.

	\subsection{Latent Space Interpolation on CelebA}
	
	To explore the latent manifold structure, we investigate the latent space interpolations between the validation set example pair $(x,y)$ and report their auto-encoded version $G_{\psi}(E_{\theta}(x))$. We linearly interpolate between $z_x = E_{\theta}(x)$ and $z_y=E_{\theta}(y)$ with equal steps in the latent space and show the decoded images. We observe smooth transitions between the pairs of examples, and intermediate images remain plausible and realistic as shown in Figure \ref{inte}. Learning disentangled latent features is an important computer vision topic, which learns the latents codes to represent different attributes of the observations \cite{tran2017DRGAN,zhang2017age,donahue2016adversarial,zhang2017cross} and employ the disentangled features for subtasks, such as recognition, detection and cross domain generation. To demonstrate the capability of learning disentangled latent codes, we cluster the learned latent codes according to the images' attributes. the reconstructed samples with manipulated latent codes based on the attributes are listed in Figure \ref{attributes}. The results demonstrate that AAAE can achieve reliable geometry of latent space without any class information at the training stage, rather than concentrating its reconstruction mass exclusively around the training examples.

	\begin{figure}[t!]
		\centering
		\begin{subfigure}[t]{0.15\textwidth}
			\centering
			\includegraphics[height=2.5cm]{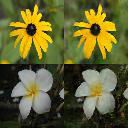}
			\caption{\label{flowers_recon_AAAE} AAAE}
		\end{subfigure}
		\begin{subfigure}[t]{0.15\textwidth}
			\centering
			\includegraphics[height=2.5cm]{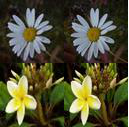}
			\caption{\label{flowers_recon_WAE} WAE}
		\end{subfigure}
		\begin{subfigure}[t]{0.15\textwidth}
			\centering
			\includegraphics[height=2.5cm]{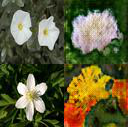}
			\caption{\label{flowers_recon_ALI} ALI}
		\end{subfigure}
		\caption{\label{flowers} Reconstructions generated by AAAE, WAE \cite{tolstikhin2017wasserstein} and ALI \cite{dumoulin2016adversarially} on the Oxford-102 dataset. (in each block, odd columns are for ground-truth, and even columns are reconstructions.)}
	\vskip -0.1in
	\end{figure}		
		\begin{figure}[t!]
		\begin{subfigure}[t]{0.15\textwidth}
			\centering
			\includegraphics[height=2.5cm]{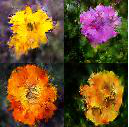}
			\caption{\label{flowers_AAAE} AAAE }
		\end{subfigure}
		\begin{subfigure}[t]{0.15\textwidth}
			\centering
			\includegraphics[height=2.5cm]{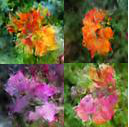}
			\caption{\label{flowers_WAE}  WAE }
		\end{subfigure}
		\begin{subfigure}[t]{0.15\textwidth}
			\centering
			\includegraphics[height=2.5cm]{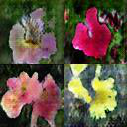}
			\caption{\label{flowers_ALI}  ALI }
		\end{subfigure}
		\caption{\label{flowers} Samples generated by AAAE, WAE and ALI on the Oxford-102 dataset.}
	\end{figure}
	\subsection{Experiments on Oxford-102}
	We finally evaluate the proposed model on the Oxford-102 dataset. The images in this dataset have large scale, pose, and light variations, which make it difficult to match the distribution $p(x)$. In this experiment, we utilize the same architectures to approximate the implicit distribution and sample from the learned distribution. As listed in Table \ref{t1}, the higher FID scores produced by each approach indicates the complexity of the dataset. Our model achieves a comparative score of 103.46 comparing to 80.817 from ALI and 196.56 from WAE. As shown in Figure \ref{flowers_recon_AAAE}, the AAAE samples capture most of the characteristic. Although ALI generates smooth samples with better FID score, they are not realistic as shown in Figure \ref{flowers_recon_AAAE}. In terms of reconstruction quality, our model is trained to generate faithful reconstructions. Some random samples are also shown in Figure \ref{flowers}. The reconstructions from the AAAE and WAE models exactly reflect the characteristics of the inputs while the reconstructions from the ALI model are far away from the inputs. 
	
	\section{Conclusion} \label{conclusion}
	In this paper, we have presented an adversarially approximated autoencoder, which learns the latent code space a manifold structure and generates high quality samples given latent codes. This is fulfilled by training an autoencoder with more flexibility without penalizing the discrepancy between the latent space distribution and a prior explicit distribution. The proposed AAAE model handles the data distribution and latent code separately. We integrate the reconstruction loss with GAN loss to enforce the match between the learned data distribution and the true distribution, and the space of latent codes is approximated by another generator with the adversarial learning. It turns out that the dual discriminators produce desired ratio estimation of the distributions without compromising the training stability. Experimental results on three datasets, including the MNIST, Cifar-10 and CelebA, show that the images sampled from the learned distribution are of better quality while the reconstructions are consistent with the datasets. As illuminated by the evaluation results on Oxford-102 dataset, the samplings are not realistic while the reconstructions are faithful to the real input data. In the future, we will extend the approximator and provide stronger regularization to enhance the match between the encoded and generated latent code distribution.  
	


	%
	%
\clearpage
\appendices
	
	\section{$\hspace{0.2cm}$ Proof of the Lower Bound}
	This appendix completes the proof of the bound that
	\begin{align}
	\begin{aligned}
	-\int p_{\phi}(c_g|z)\log p_{\theta}(c|x)dc_g &\leq KL[q_{\psi}(x|c)||p(x)] \\&+ KL[p_{\phi}(c_g|z)||p_{\theta}(c|x)] \\&-\mathbb{E}[\log p_{\phi}(c_g|z)]
	\end{aligned}
	\end{align}
	
	Starting from $-\log p_{\theta}(c|x) =  -\log\int p_{\theta}(c|x)p(x) dx$, introducing a constant yields
	\begin{align}
	\begin{aligned}
	-\log p_{\theta}(c|x) &=  -\log\int p_{\theta}(c|x)p(x)\frac{q_{\psi}(x|c)p_{\phi}(c_g|z)}{q_{\psi}(x|c)p_{\phi}(c_g|z)}dx \\
	&\leq   -\int q_{\psi}(x|c)\log\frac{p_{\theta}(c|x)p(x)}{q_{\psi}(x|c)p_{\phi}(c_g|z)} dx \\&- \log p_{\phi}(c_g|z) \\
	&=   \int q_{\psi}(x|c)\log\frac{q_{\psi}(x|c)p_{\phi}(c_g|z)}{p_{\theta}(c|x)p(x)} dx\\& - \log p_{\phi}(c_g|z)
	\end{aligned}
	\end{align}
	
	Thus, we obtain the final objective function
	\begin{align}
	\begin{aligned}
	H&(G_{\phi},p_{\theta})  = -\int p_{\phi}(c_g|z)\log p_{\theta}(c|x)dc_g\\
	&\leq  \int \int p_{\phi}(c_g|z)q_{\psi}(x|c)\log\frac{q_{\psi}(x|c)p_{\phi}(c_g|z)}{p_{\theta}(c|x)p(x)} dxdc_g\\& - \int p_{\phi}(c_g|z)\log p_{\phi}(c_g|z)dc_g \\
	& =  \int \int p_{\phi}(c_g|z)q_{\psi}(x|c)\bigg(\log\frac{q_{\psi}(x|c)}{p(x)} +\log\frac{p_{\phi}(c_g|z)}{p_{\theta}(c|x)}\bigg) dxdc_g \\&- \int p_{\phi}(c_g|z)\log p_{\phi}(c_g|z)dc_g \\
	& =  \int  q_{\psi}(x|c)\log\frac{q_{\psi}(x|c)}{p(x)} dx+\int p_{\phi}(c_g|z) \log\frac{p_{\phi}(c_g|z)}{p_{\theta}(c|x)} dc_g \\&- \int p_{\phi}(c_g|z)\log p_{\phi}(c_g|z)dc_g \\
	&= KL[q_{\psi}(x|c)||p(x)] + KL[p_{\phi}(c_g|z)||p_{\theta}(c|x)] \\&-\mathbb{E}[\log p_{\phi}(c_g|z)]
	\end{aligned}
	\end{align}
	This completes the proof.
	\vspace{5cm}
	
	\section{$\hspace{0.2cm}$ Hyperparameters}
	The model architecture and hyperparameters are shown in Table \ref{tab:MNIST_description}. We set $z\in \mathbb{R}^{64}$ and $c\in \mathbb{R}^{128}$. Note that the experiments on other datasets are performed by either adding or deleting the convolution layers according to the image resolution.
	\vspace{0.5cm}
		
	\begin{table}[h!]
		\caption{\label{tab:MNIST_description} MNIST model hyperparameters (unsupervised).}
		\begin{tabular}{@{}rllcp{0.1cm}c@{}} \toprule
			Operation              & Kernel       & Strides      &\#output & BN?      &  Activation \\ \midrule
			\multicolumn{6}{@{}l@{}}{	$E_{\theta}(c|x)$: $3 \times 32 \times 32$   }                                                           \\
			Convolution            & $4 \times 4$ & $1 \times 1$ & $64$         & $\surd$       & eLU \\
			Convolution            & $4 \times 4$ & $2 \times 2$ & $128$         & $\surd$       & eLU \\
			Convolution            & $4 \times 4$ & $2 \times 2$ & $256$        & $\surd$      & eLU \\
			Convolution            & $4 \times 4$ & $2 \times 2$ & $512$        & $\surd$      & eLU \\
			Convolution            & $4 \times 4$ & $2 \times 2$ & $1024$        & $\surd$      & eLU \\\
			Fully Connected        &  &  & $128$        & $\times$    & \\
			\hline 
			\multicolumn{6}{@{}l@{}}{$G_{\psi}(x|c)$: $128$  }                                                             \\
			Fully Connected		   &  && 1024        & $\surd$       & eLU \\
			Transposed convolution & $4 \times 4$ & $2 \times 2$ & $512$        & $\surd$       & eLU \\
			Transposed convolution & $4 \times 4$ & $2 \times 2$ & $256$         & $\surd$      & eLU \\
			Transposed convolution & $4 \times 4$ & $2 \times 2$ & $128$         & $\surd$      & eLU \\
			Transposed convolution & $4 \times 4$ & $2 \times 2$ & $64$         & $\surd$     & eLU \\
			Convolution            & $3 \times 3$ & $1 \times 1$ & $3$          & $\times$     & Tanh    \\
			\hline
			\multicolumn{6}{@{}l@{}}{$D_{\omega}(x)$: $3 \times 32 \times 32$ }                                                              \\
			Convolution            & $4 \times 4$ & $2 \times 2$ & $64$         & $\times$      &Leaky ReLU \\
			Convolution            & $4 \times 4$ & $2 \times 2$ & $128$         & $\times$      &Leaky ReLU  \\
			Convolution            & $4 \times 4$ & $2 \times 2$ & $256$        & $\times$      &Leaky ReLU    \\
			Convolution            & $4 \times 4$ & $2 \times 2$ & $512$        & $\times$      &Leaky ReLU     \\
			Convolution            & $3 \times 3$ & $1 \times 1$ & $1$        & $\times$    &    \\
			\hline
			\multicolumn{6}{@{}l@{}}{$G_{\phi}(c|z)$: $64$ }                                                                \\
			Fully Connected            & &  & $512$       & $\times$     &eLU       \\
			Fully Connected            &  &  & $512$       & $\times$   &eLU       \\
			Fully Connected            &  &  & $512$       & $\times$   &eLU       \\
			Fully Connected            &  &  & $128$       & $\times$   &       \\ 
			\hline
			\multicolumn{6}{@{}l@{}}{$D_{\gamma}(c)$: $128$  }                                                          \\
			Fully Connected            & &  & $512$       & $\times$     &Leaky ReLU       \\
			Fully Connected            &  &  & $512$       & $\times$   &Leaky ReLU       \\
			Fully Connected            &  &  & $512$       & $\times$   &Leaky ReLU       \\
			Fully Connected            &  &  & $1$          & $\times$     &    \\ \midrule
			\multicolumn{6}{@{}l@{}}{Optimizer Adam: ($\alpha = 10^{-4}$, $\beta_1 = 0.5$, $\beta_2 = 10^{-3}$)} \\
			\multicolumn{6}{@{}l@{}}{Batch size: 100}												  \\
			\multicolumn{6}{@{}l@{}}{Epochs: 200}  											  \\
			\multicolumn{6}{@{}l@{}}{Leaky ReLU slope, maxout pieces: 0.1, 2}                                                \\
			\multicolumn{6}{@{}l@{}}{Weight, bias initialization: Isotropic gaussian ($\mu = 0$, $\sigma = 0.01$), Constant($0$)} \\ \bottomrule
		\end{tabular}
	\end{table}
	
	\section{ $\hspace{0.2cm}$  Additional Results}
	Some additional experimental results are listed in this section.
	\begin{figure*}[h]
		\begin{center}
			{\includegraphics[height=9cm]{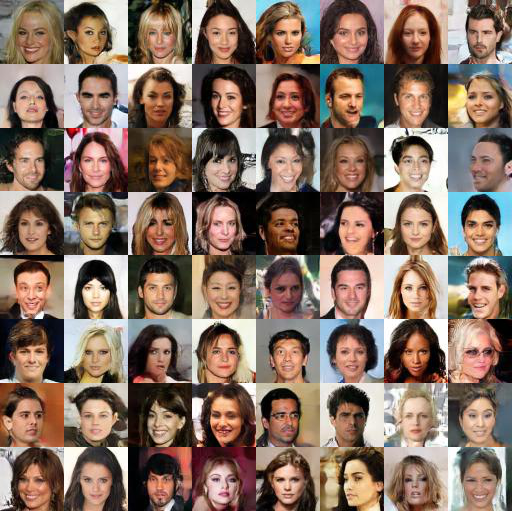}}
		\end{center}
		\caption{Generated samples trained on the CelebA dataset.
		}
		\label{f3}
	\end{figure*}
	\begin{figure*}[h]
		\begin{center}
			{\includegraphics[height=10cm]{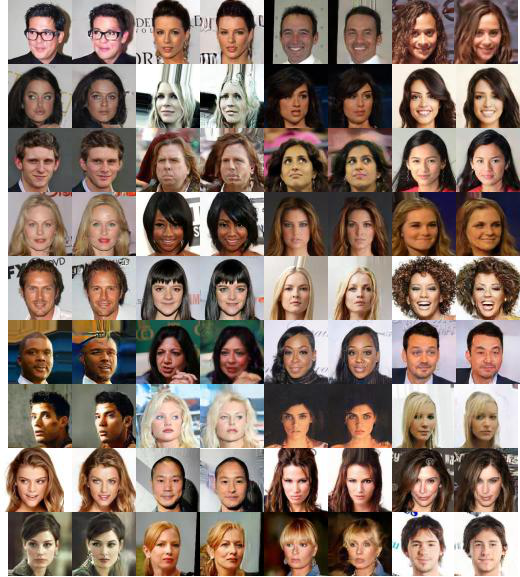}}
		\end{center}
		\caption{Generated reconstructions trained on CelebA: Odd columns stand for ground-truth, and even columns for reconstruction.}
		\label{f3}
	\end{figure*}
	
	\begin{figure*}[h!]
			\centering
		\begin{subfigure}[h]{0.3\textwidth}
			\includegraphics[height=2.5cm]{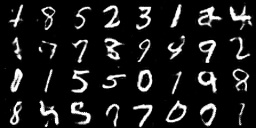}
			\caption{\label{fig:cifar} MNIST samples}
		\end{subfigure}
		\begin{subfigure}[h]{0.3\textwidth}
			\includegraphics[height=2.5cm]{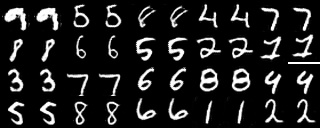}
			\caption{\label{fig:cifar_cosntrcution} MINST reconstructions}
		\end{subfigure}
		\caption{\label{fig:cifar_construction} Generated samples and reconstructions trained on the MNIST dataset}	
			\vskip -0.2in
	\end{figure*}	
	\begin{figure*}[h!]
		\begin{center}		
			\includegraphics[width=0.89\textwidth]{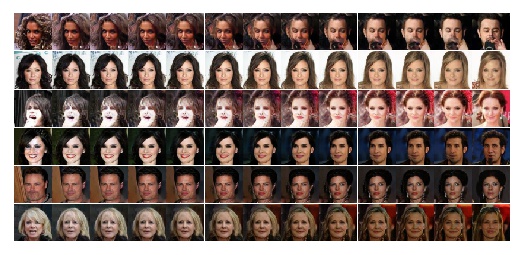}
		\end{center}
		\caption{Latent space interpolations on the CelebA validation set. Left and right column are the original data pairs. The columns in between are the reconstructions generated with linearly interpolated latent codes.}
		\label{f3}
	\end{figure*}
	
\clearpage	
\balance
	\bibliographystyle{IEEEtran}
	\bibliography{AAAE_arxiv}
\end{document}